%% file: halpern_mlhc.tex
\pdfoutput=1
\documentclass[twoside,11pt]{article}

\usepackage{mlhc}
\usepackage{natbib}
\usepackage{todonotes}
\usepackage{amsmath,comment}
\usepackage{algorithm}
\usepackage{algorithmicx}
\usepackage{todonotes}

\DeclareMathOperator*{\argmin}{arg\,min}

\ShortHeadings{Clinical Tagging with Joint Probabilistic
  Models}{Halpern, Horng, MD, and Sontag, PhD}

\begin{document}

\title{Clinical Tagging with Joint Probabilistic Models}

\author{\name Yoni Halpern \email halpern@cs.nyu.edu \\
       \addr Department of Computer Science\\
       New York University\\
       New York, NY, USA
       \AND
       \name Steven Horng \email shorng@bidmc.harvard.edu \\
       \addr Department of Emergency Informatics\\
       Beth Israel Deaconess Medical Center\\
       Boston, MA, USA
       \AND
      \name David Sontag \email dsontag@cs.nyu.edu \\
      \addr Department of Computer Science\\
      New York University\\
      New York, NY, USA} 
\maketitle

\input{abstract}
\input{introduction}
\input{methods}

\input{results}

\input{discussion}
\section*{Acknowledgments}
This work is partially supported by a Google Faculty Research Award 
and NSF CAREER award \#1350965. Y.H. was supported by a postgraduate scholarship
from the Natural Sciences and Engineering Research Council of
Canada (NSERC). 
\bibliography{uai16}
\appendix
\input{appendix}

\end{document}

%% file: abstract.tex
\begin{abstract}
We describe a method for parameter estimation in bipartite probabilistic graphical models for joint prediction of clinical conditions from the electronic medical record.  The method does not rely on the availability of gold-standard labels, but rather uses noisy labels, called anchors, for learning.
We provide a likelihood-based objective and a moments-based initialization that are effective at learning the model parameters. 
The learned model is evaluated in a task of assigning a heldout clinical condition to patients based on retrospective analysis of the records, and outperforms baselines which do not account for the noisiness in the labels or do not model the conditions jointly.
\end{abstract}

%% file: introduction.tex

\section{Introduction}

Clinical decision support systems aim to relay clinically relevant information while the patient is being treated.
The relevant information can vary and can include recommendations of standardized pathways of care \citep{panella2003reducing}, evidence-based guidelines, and warnings about allergies and other contraindicated medications.
The most effective systems are those that can {\em understand} the patient's electronic medical record as it is being populated.
By harnessing information entered as part of the clinician's regular workflow, these systems do not add additional work or cognitive burden and integrate seamlessly into clinical care. 

While the field of machine learning has shown tremendous success learning to recognize patterns from large collections of labeled examples, one issue that arises repeatedly when applying machine learning to medical applications is the difficulty and cost of obtaining accurate labels for training. 
In this work, we focus on the so-called ``anchored'' setting, where gold-standard labels are difficult to obtain, but noisy versions of these labels can be easily extracted from clinical text using simple rules~\citep{HalpernEtAl_amia14, agarwal2016learning}. These rules (called anchors) are then used as surrogate labels in standard machine learning pipelines, with appropriate adjustments to account for noise~\citep{natarajan2013learning, elkan2008learning}. Anchors need to be specified manually by experts, but are much easier than labeling large numbers of patients with manual chart abstraction.

Previous work with anchors~\citep{halpern2016electronic} showed that they can be used to build a large number of individual classifiers to identify a range of clinical conditions, but did not address the joint modeling of these conditions. Without a joint model, it is difficult to calibrate the individual classifiers against each other to properly respond to queries such as: ``what are the most likely clinical conditions for this patient?'' or ``what {\em else} might the patient have?''

In this work, we present a method of training joint probabilistic models with anchors, and show an improvement of the joint model over individual classifiers in a clinical condition tagging task, specifically in answering the question ``what else might the patient have?'' 
\vspace*{-1em}

\section{Modeling clinical conditions}
\label{sec:condition}
We model 23 clinical conditions relevant to the emergency department. 
These are a subset of the conditions modeled in~\cite{halpern2016electronic}, chosen to have anchors which include both ICD9 billing codes and another form of observation (either free text or medication). The anchors are available at \url{https://github.com/clinicalml/clinical-anchors}. To simulate the anchor setting while still having labels for evaluation, we use the ICD9 codes to represent  ``ground truth'' for the purposes of this study and other observations such as medications or free-text for anchors and features. 
While ICD9 codes are generally unreliable for establishing gold-standard clinical conditions~\citep[e.g.,][]{cipparone2015inaccuracy, tieder2011accuracy, birman2005accuracy, aronsky2005accuracy}, we consider them reliable enough to assess relative performance of different methods which are trained using anchors. Table~\ref{tab:conditions} gives the full list of clinical conditions that were modeled.

\begin{table}[tb]
  \centering
{\small
  \begin{tabular}{|lll|}
  \hline
\multicolumn{3}{| c |}{{\bf Clinical Conditions}}\\
  \hline
abdominal pain acute &  alcohol acute & allergic reaction acute\\
asthma-copd acute &  back pain acute &  cellulitis acute\\
cva acute & epistaxis acute & fall acute\\
gi bleed acute & headache acute & hematuria acute\\
intracranial hemorrhage acute &  kidney stone acute & vehicle collision acute\\
pneumonia acute &  severe sepsis acute & sexual assault acute\\
suicidal ideation acute & syncope acute & uti acute\\
liver history & hiv history & \\
  \hline
  \end{tabular}
 }
\caption{\small \label{tab:conditions}The full list of clinical conditions included in the model. Acute conditions relate to the patient's current condition. History refers to the patient's history.}
\vspace*{-4ex}
\end{table}

%% file: methods.tex

\section{Cohort}

The study was performed in a 55,000-visit/year level 1 trauma center and tertiary  academic  teaching  hospital.  
All consecutive  emergency  department (ED) patients between 2008 and 2013 were included.
Records were de-identified~\citep{neamatullah2008automated} and personal health information was removed before beginning the analysis. 
Each record represents a single patient visit, leading to a total of 273,174 records of emergency department patient visits.
The study was approved by the hospital's institutional review board.

\subsection{Cohort Selection} 
We focus on patients with at least two of the modeled clinical conditions so that it is possible to identify one condition and ask ``what else might the patient have?''
After filtering for patients with at least two of the modeled conditions, we were left with 16,268 patients.
Of these patients, 11,000 were designated for training and 5,000 for testing. The final 268 patients were not used.

\subsection{Data extraction and feature selection}
\label{features}
For each visit, we extracted data from the fields listed in Table~\ref{tab:features} to build a binary bag-of-words representation for every patient. Full details of the free-text processing pipeline including negation and bigram detection can be found in Appendix~\ref{app:text}. For each condition, we create an anchor token which appears if {\em any} of the condition's anchors appears in the record. Terms that appear in more than 50\% of the patient records are removed as stopwords, and the most common 1000 terms are kept. Any anchors that were filtered out in this step are added back in, yielding a final feature vector with 1003 binary indicators.

\begin{table}[t]
  \centering
{\small
  \begin{tabular}{|l|l|}
  \hline
  \textbf{Field} & \textbf{Representation} \\
  \hline
  Age & Binned to nearest decade \\
  Sex & M / F \\
  Chief Complaint & Free text \\
  Triage Assessment & Free text \\
  MD comments & Free text \\ 
  Medication history & GSN codes \\
  Dispensed medications & GSN codes \\
  Billing codes & ICD9 codes \\
  \hline
  \end{tabular}
}
  \caption{\small \label{tab:features} Features extracted. Billing codes were extracted to perform the evaluation, but were not used to create the patient feature vector.}
  \vspace{-5ex}
\end{table}

\section{Methods}

We model conditions and observations as a bipartite Bayesian network. In the following sections, we will describe the structure of the model and methods for learning its parameters. Throughout, we will follow the convention that random variables are denoted by uppercase letters (e.g., $Y_i$) and their values indicated by lowercase variables ($y_i \in \{0,1\}$).

\subsection{Anchor assumption}
\label{sec:anchor_assumption}

We assume the anchors are corrupted versions of the true labels and that the corruption process obeys a conditional independence constraint:  The state of the anchor depends only on the true label. Specifically, conditioned on the true label, it is independent of all other observations. \citet{HalpernEtAl_amia14} additionally required a positive-only assumption (the corruption process does not produce false positive cases), which we do not require here. Instead we will require that that the class-conditional noise rates of the corruption process are known.

\subsection{Model structure}
\begin{figure}[tb]
  \centering
  \includegraphics[scale=0.53]{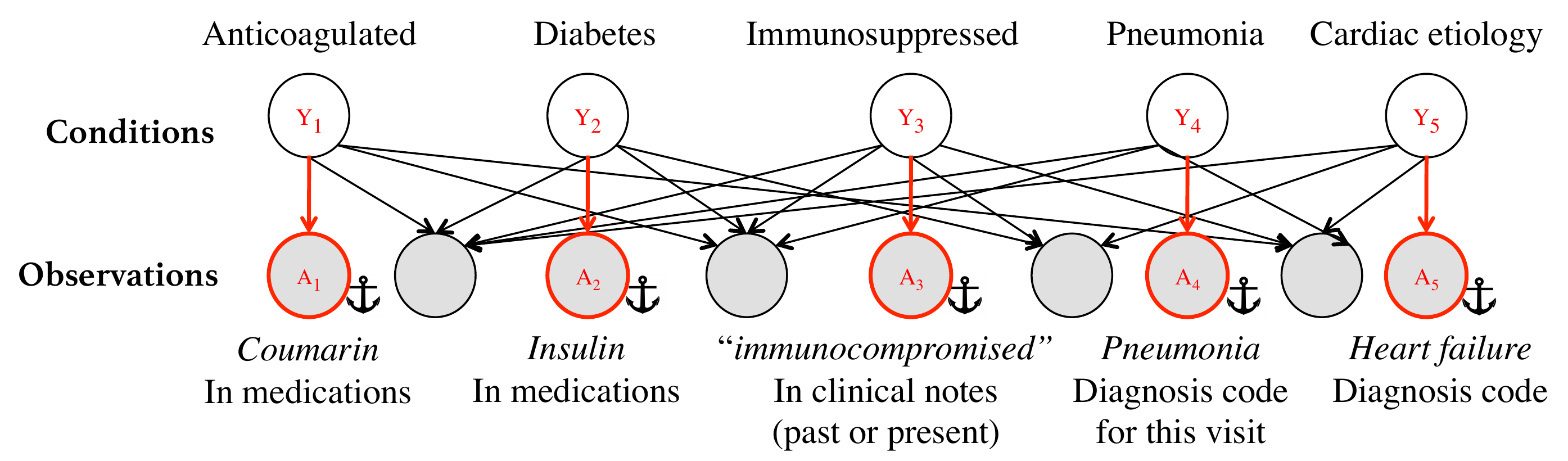}
  \vspace*{-4mm}
  \caption{\small The joint probabilistic model used for clinical tagging is a bipartite graph involving conditions and observations, like QMR-DT \citep{QMR_DT} with one or more {\em anchor} (red outline) for each condition (only one anchor per condition is shown in the illustration). Other observations (black outline) can have multiple parents.}
  \label{fig:QMR-DT}
  \vspace*{-4mm}
\end{figure}

We use a graphical model patterned after the historical QMR-DT network for diagnosis~\citep{QMR_DT} (see Figure~\ref{fig:QMR-DT}). The Bayesian network consists entirely of binary random variables, which are partitioned into conditions $(Y_1, ..., Y_m)$ and observations $(X_1, ..., X_n)$. The model is bipartite with directed edges from conditions to observations. 

The conditions are assumed to be marginally independent with individual prior probabilities, denoted as $\pi_i$:
\vspace{-3mm}
\begin{equation}
P(Y=\{y_1, ..., y_m\}) = \prod_{i=1}^m \pi_i^{y_i}(1-\pi_i)^{1-y_i}.
\end{equation}
\vspace{-2mm}

The conditional probabilities of the observations (given the state of the conditions), are parametrized with a ``noisy-or'' distribution (Equation~\ref{eq:noisy_or}) \citep{QMR_DT,pearl1988probabilistic}:
\begin{equation}
\label{eq:noisy_or}
P(X_j=0 | Y=\{y_1, ..., y_m\}) = (1-l_{j})\prod_{i=1}^m f_{i,j}^{y_i},
\end{equation}
where the parameters $f_{i,j}$ are referred to as {\em failure} probabilities and $l_j$ is the {\em leak} probability. 

The network can be viewed as a generative model. For each new patient, each condition $Y_i$ independently turns on with probability $\pi_i$. Each condition which is on tries to turn on each of its children, $X_j$, but fails with probability $f_{i,j}$. An additional ``noise'' parent is always on and fails to turn on its children with probability $(1-l_j)$.

Rather than naively treat the anchors (which are unreliable labels) as telling us which conditions are present, we treat the conditions as latent variables and treat the anchors as observations. 
The anchor assumption places a structural constraint on the graphical model. 
Specifically, each anchor $A_i$ may only have one parent which is $Y_i$.

Treating the conditions as latent variables makes learning the parameters of the model computationally difficult. When the variables are all observed, maximum likelihood estimation of the parameters is a concave optimization problem that can be solved efficiently. However, when the conditions are unobserved, the problem is no longer concave and optimization procedures get stuck in local minima, even when the anchor assumption holds true. 
In the following two sections, we describe a likelihood-based objective and an effective initialization that allow us to learn models with good predictive capabilities.

\subsection{Semi-supervised objective}
\label{sec:semisupervised}

Exact inference in the QMR-DT model is known to be NP-hard, even if there is an anchor for every condition. The typical approach maximizing likelihood in the presence of latent variables, Expectation Maximization (EM)~\citep{DLR}, is computationally challenging because at each step one has to use an approximate inference algorithm such as Markov chain Monte Carlo to approximate the necessary expectations.

Instead, we follow \citet{mnih2014neural} in formulating a variational lower bound on the likelihood function using a recognition model. We start with the standard Evidence Lower BOund (ELBO), which holds true for {\em any} distribution, $q(Y|X)$:
\begin{equation}
\mathcal{L}(\theta, q) \equiv E_{y \sim q}\left[\log P(X,y; \theta) - \log q(y|X)\right] \leq \log P(X).
\end{equation}

In mean field variational inference, $q$ is chosen to be a fully factorized distribution. In this work, we restrict $q$ to be the output of a parametrized model, with parameters $\phi$. The parametrized distribution $q(y|X; \phi)$ is referred to as a {\em recognition} model and its function is to perform approximate inference in the network. The bound is tightest as $q(Y|X)$ approaches $P(Y|X; \theta)$, that is, as it approximates inference in the generative model. As learning proceeds, the recognition model learns to {\em compile} inference. 

In our work we use a simple recognition model that performs logistic regression to approximate the posterior of each condition independently:
\vspace{-2mm}
\begin{equation}
q(y | x; \phi) = \prod_i \Big(y_i \sigma(\phi_i \cdot x) + (1-y_i)(1-\sigma(\phi_i \cdot x))\Big),\vspace{-2mm}
\end{equation}
where $\sigma$ is the sigmoid function, $\sigma(x) = \frac{1}{1+e^{-x}}$ and $x$ is padded with a 1 to allow for a bias term. In contrast with mean-field inference, where $q$ is optimized for every data point separately, here we train a single model, allowing us to amortize the cost of inference over many data points. Methods to take gradients with respect to $\theta$ and $\phi$ and optimize with stochastic gradient ascent are described in detail in~\citet{mnih2014neural}. Our hyperparameter settings are described in Appendix~\ref{app:parameters}. Using this algorithm enabled us to learn orders of magnitude faster than EM, with comparable results in terms of likelihood objective obtained. 

However, we found that without adding an additional term to the objective, the anchor constraints (i.e., each anchor only has a single, specified parent) were not sufficient to make sure that the latent variables took on their intended meanings. Specifically, we observed that as the held-out likelihood of the observations improved, the predictive quality of the models got worse (measured using the heldout tag prediction task of Section~\ref{sec:heldout_tag}). Upon inspecting the model, we found that drift occurred: the latent variables took on new meanings and lost their original grounding.

Inspired by recent work on semi-supervised learning with deep generative models \citep{KingmaEtAl_NIPS14}, our solution is to add a supervised term to the objective that encourages the recognition model to additionally predict the presence or absence of the anchors, ensuring that the meaning of the latent variable is tightly tied to the anchor. 
The prediction cannot use the anchors themselves, so we form a new censored vector, $\tilde{x}$, which is a copy of $x$ but has the values of the anchors set to a constant 0. We also introduce an additional bias term $\phi_0^\prime$ to allow the prediction of the anchors and the labels to differ from each other. The supervised term has the form:
\begin{equation}
R(\phi, \phi_0^\prime) = -\ell(\sigma(\phi \cdot \tilde{x} + \phi_0^\prime), a),
\end{equation} 
where $\ell(\cdot, \cdot)$ is log loss and $a$ is the vector of anchors. The final objective is thus:
\begin{equation}
\label{eq:final_obj}
\mbox{maximize  } \mathcal{L}(\theta, \phi) + \lambda R(\phi, \phi_0^\prime),
\end{equation}
where $\lambda>0$ is a hyperparameter specifying the trade-off between these two terms in the objective, and although we wrote Eq.~\ref{eq:final_obj} for a single data point, the actual objective we optimize is the sum of this over all the data points. A more detailed version of the objective is found in Appendix~\ref{app:semisupervised}.

At test time, we discard the recognition model, $q_\phi$, which was used to train the parameters of the generative bipartite Bayesian network, but cannot support queries with conditioning on some of the clinical variables, and use the joint probabilistic model, $P(X,Y; \theta)$, for inference. Inference in the joint model does not have an efficient closed form solution, but can be approximated with Gibbs sampling. 

\subsection{Model initialization}
\label{sec:init}
In order to initialize the model, we use the anchors to get a rough estimate of the failure probabilities for each of the observations. If we observed the latent clinical conditions (the $Y$ variables), we could use a simple moments-based estimator using empirical counts to estimate the failure probabilities (Equation~\ref{eq:moments-estimator}). 

\begin{equation}
\label{eq:moments-estimator}
\hat{f}_{i,j} = \frac{\hat{P}(X_j=0 | Y_i=1)}{\hat{P}(X_j=0 | Y_i=0)}.
\end{equation}

The estimator $\hat{f}_{i,j}$ is then clipped to lie between [0,1]. The consistency of the method is not affected by this clipping, since if sufficient data were drawn from the model, the estimator would naturally lie in that range and clipping would not be necessary. Once all the failure probabilities are estimated, the leak probabilities can be estimated to account for the difference between the true observed counts and those predicted by the model (Appendix~\ref{app:leak}).

Since the clinical conditions are generally unobserved, we estimate these conditional probabilities using empirical counts assuming that the anchors (which are noisy versions of the labels) are $Y$, and then perform a {\em denoising} step to estimate the conditional probabilities as though the true labels were observed. Specifically, in Section~\ref{sec:anchor_assumption} we assumed that the label corruption process was independent of all other observed variables. This leads to the following equation:
\begin{equation}
P(X_j | A_i) = P(Y_i = 1 | A_i) P(X_j | Y_i=1) + P(Y_i = 0 | A_i) P(X_j | Y_i=0)
\end{equation}

The left-hand side of this equation is a quantity that only involves observed variables $(X_j, A_i)$ and can be estimated from empirical counts. The right-hand side uses the noise rates of the corruption process $P(Y_i|A_i)$ and the conditional probabilities that we care about, $P(X_j |Y_i)$. If we assume that the noise rates of the corruption process are known, then we can form four independent linear equations with four unknowns and solve the following matrix equation:
\begin{equation}
\label{eq:matrix}
\vec{P}(X_j | A_i) = R \vec{P}(X_j | Y_i),
\end{equation}
where $\vec{P}(X_j | A_i)$ is a column vector with four entries, one for each setting of $(X_j, A_i)$ in $\{0,1\}^2$. $R$ is a $4 \times 4$ matrix encoding the noise rates of the corruption process. Explicit constructions of these terms are given in Appendix~\ref{app:matrix}.

We could simply invert the noise matrix $R$ to solve $\vec{P}(X_j | Y_i) = R^{-1} \vec{P}(X_j | A_i)$, however, it would not be guaranteed that the solution would give a valid probability (i.e., non-negative and sum-to-one conditions) for $\vec{P}(X_j | Y_i)$. Instead, we explicitly solve the optimization problem with simplex constraints to minimize a KL-divergence measure between a proposed distribution $\vec{P}(X_j | Y_i)$ and the denoised version of the empirical counts $P(X_j | A_i)$:

\begin{align}
\label{eq:optimization}
\vec{P}(X_j | Y_i) =& \argmin_{\vec{p} \in \Delta} \text{D}_{\text{KL}} \left(\vec{P}(X_j | A_i) \middle|\middle| R \vec{p}\right)
\end{align}

The optimization is convex and we solve it with exponentiated gradient descent~\citep{Kivinen95exponentiatedgradient}. The {\em cleaned} distribution, $\vec{P}(X_j | Y_i)$, obtained from solving Equation~\ref{eq:optimization} is then substituted into the failure probability estimator in Equation~\ref{eq:moments-estimator} to obtain estimates of the failure and leak probabilities (regarding the leak probabilities, see Appendix~\ref{app:leak}). This whole procedure can be shown to be a {\em consistent} estimator, meaning that if the model assumptions hold (i.e., of conditional independence), this will converge to the true probabilities as the amount of data goes to infinity.

\begin{algorithm}[tb]
\caption{Parameter estimation algorithm}
\label{alg:full_story}
\begin{algorithmic}[1]
\State (Precondition) Identify anchors
\State Obtain cleaned moment estimates using anchors (Eq.~\ref{eq:optimization})
\State Initialize $\theta_{0}$ using method of moments (Eq.~\ref{eq:moments-estimator}).
\State Initialize $\phi_0$ randomly
\State NVIL optimization \citep{mnih2014neural} of Eq.~\ref{eq:final_obj}.
\State Discard $\phi$ and use joint model parametrized by $\theta$.
\end{algorithmic}
\end{algorithm}

\subsection{Complete algorithm}
The full parameter estimation algorithm is summarized in Algorithm~\ref{alg:full_story}.

\subsection{Model selection}
\label{sec:model_selection}
We do not assume that we have any ground truth labels to do model selection, so we use a stopping criteria based on heldout {\em anchor} prediction. 
We hold out 1000 patients from the train set as a validate set to determine the stopping criteria.
For each patient in the validate set, we censor one positive anchor that appears in the record and all of the negative anchors, and perform inference with the joint model $P(X,Y; \theta)$ to determine which anchor is missing. Inference for the final anchor is performed with Gibbs sampling to obtain marginals for the tags, and then the likelihood of each anchor is computed as a function of the marginal likelihood of its parent tag (details in Appendix~\ref{app:lastanchor}). This model selection criterion mimics the heldout-tag prediction task described in Section~\ref{sec:heldout_tag} used in the evaluation, but uses anchors instead of the true labels. Other stopping criteria using anchors could be developed depending on the intended use.

%% file: results.tex
\section{Results} 
\subsection{Heldout tag prediction task} 
\label{sec:heldout_tag}
We test the ability of our model to perform inference by presenting it with a heldout-tag prediction task. 
Each clinical state is assigned a tag. The model is presented with a patient record and all but one of the tags that apply for that record. 
The task is to predict the final tag that applies for this patient. We record the accuracy of each model (i.e., the proportion of times it chooses the correct tag to fill in), top-5 performance (i.e., proportion of times the correct tag appears in the top 5 predictions) and the mean-reciprocal rank  of the correct prediction (MRR). Since this task is choosing the best last tag, we do not need to perform approximate inference. Instead we evaluate the likelihood of each possible final tag and perform the normalization, which corresponds to exact inference (Appendix~\ref{app:lasttag}). 

In addition to highlighting the joint modeling aspect of the new model, that it is able to respond to arbitrary inference queries with conditioning, this task is also clinically relevant in that it can be used to combat a recognized cognitive bias known as ``search satisfaction error''~\citep{groopman2007doctors}. This cognitive bias, a failing of the Occam's razor heuristic, is the tendency to overlook additional conditions once a single unifying diagnosis is found. This is particularly dangerous when a more serious diagnosis is overlooked because a unifying diagnosis was discovered first. For example, a patient with signs of a severe infection may be diagnosed with urinary tract infection and treated with antibiotics, while missing a second diagnosis of pneumonia. The delay in treatment of the pneumonia could be potentially dangerous to the patient. Another example is a patient with a kidney stone, whose co-existing urinary tract infection was missed. Although kidney stones generally resolve on their own, patients who also have a concurrent urinary tract infection require immediate intervention.

Clinical decision support systems can mitigate this cognitive bias by suggesting additional diagnosis that may explain a set of symptoms. We simulate this problem by randomly removing a tag, and trying to recover it using the model. This could correspond to a situation where the physician has ``confirmed'' one diagnosis and the model attempts to suggest a likely second diagnosis that would go along with the first. Since the model performs inference, this could potentially be very different from the next most likely tag if the model received no confirmation of the first tag. 

\begin{figure}[t]
  \centering 
  \includegraphics[scale=.8]{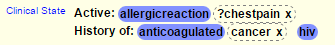}
  \caption{\small Tagging system within an electronic medical record.}
  \label{fig:tags}
\end{figure}

This tagging system is currently implemented with the electronic medical record system in the emergency department Beth Israel Deaconess Medical Center in Boston, MA. Physicians are presented with possible tags for each patient (see figure~\ref{fig:tags}).  Confirming or rejecting a tag updates recommendations through inference.

\subsection{Baselines}
\label{sec:baselines}
We compare against three different baselines. Since our proposed method assumes knowledge of the correct failure and leak parameters for the anchors, we provide that information to all of the baselines for fairness.
\begin{enumerate}
\item {\bf Naive labels} -- treats the anchors as true labels and learns a noisy-or model with maximum likelihood estimation. In the final model, we ``edit'' the failure and leak parameters of the anchors to set them to the correct values.
\item {\bf Noise tolerant classifiers} -- We use the method presented in~\citet{natarajan2013learning} to learn noise-tolerant classifiers (explicitly providing the algorithm with the noise rates). In experiments we found that this method was more effective than the method of \citet{elkan2008learning}, so we present this baseline to compare to the individual classifiers learned in~\citet{HalpernEtAl_amia14}. The method of~\cite{natarajan2013learning} does not explictly describe how to predict when anchors are observed in the record. In this case, we simply predict the noise rate of the anchor, which we found to be more effective than ignoring the special status of the anchor.
\item {\bf Oracle MLE} -- An upper bound on performance which uses the true labels to learn a a noisy-or model with maximum likelihood estimation. This is impractical in practice, but gives us a sense of how close to optimal we are performing using our noisy labels.
\end{enumerate}

\subsection{Results on heldout tag prediction} 

Table~\ref{tab:baseline_results} presents our method compared with the baselines presented in Section~\ref{sec:baselines}. `Noisy-or init' refers to the moments-based estimator described in Section~\ref{sec:init}, whereas `Noisy-or final' refers to the results after the semi-supervised learning algorithm described in Section~\ref{sec:semisupervised}.

\begin{table}[t]
  \centering 
  \begin{tabular}{|l|l|l|l|}\hline
    {\bf Model}           & {\bf Accuracy} & {\bf Top 5} & {\bf MRR} \\ 
    \hline
    Noise tolerant classifiers & 0.54         & 0.86            & 0.67\\
    Naive Labels               & 0.61         & 0.85            & 0.71\\ 
    Noisy-or init              & 0.64         & 0.91            & 0.76 \\
    {\bf Noisy-or final}       & {\bf 0.68}   & {\bf 0.92}      & {\bf 0.79}\\
\hline
    Noisy-or oracle MLE        & 0.71         & 0.93            & 0.81 \\ 
    \hline 
  \end{tabular}
  \caption{\small \label{tab:baseline_results} Results for last-tag prediction. 
  Performance measures are Accuracy, Top-5 (correct tag within the top 5) and MRR (mean reciprocal rank).
  Noisy-or init uses the model with the $\theta_0$ parameters. Noisy-or final shows the result after likelihood optimization. \vspace*{-4mm}} 
\end{table}

The noisy-or model significantly outperforms the noise-tolerant classifiers and the naive labeling baselines. Our performance comes close to the optimal maximum likelihood performance, suggesting that even though we don't use the true labels in training, we are still able to recover a model which is similar to the one we would learn if we had access to the true labels. The method of moments initialization is helpful. Using random initialization, we do not beat the naive labels baseline. Appendix~\ref{app:course} shows that the likelihood and tagging objectives are aligned after introducing the semi-supervised objective. Table~\ref{table:factors} shows the highly weighted words learned by our model. All of the noisy-or models learn similar sets of highly weighted words, the main differences between the models are in the exact settings of the parameters which affect the inference procedure to choose the correct last tag. 

\begin{table}[t]
\centering

\begin{tabular}{|c|l|}
\hline
{\bf Tag} & {\bf Top weighted terms} \\
\hline
abdominal pain & pain, Ondansetron, nausea, days \\
alcohol & male, sober, ed, admits, found, denies \\
asthma & albuterol sulfate, sob, Methylprednisolone, cough \\
fall & s/p fall, fall, fell, neg:loc, neg:head \\
hematuria & infection, male, urology, urine, flank pain \\
HIV+ & male, Truvada, cd4, age:40-50, Ritonavir \\
collision & car, neg:loc, age:20-30, hit, neck, driver \\
\hline
\end{tabular}
\caption{\small \label{table:factors}  Highly weighted (low failure probability) words learned by the noisy-or model after likelihood optimization. 
Words marked neg: are within a negation scope. Some shortforms are present in the text (ed: emergency department, sob: shortness of breath, loc: loss of consciousness, s/p: status post).}\vspace{-4mm}
\end{table}

%% file: discussion.tex
\section{Discussion} 

There are a number of limitations to this study. First, we used the ED ICD-9-CM discharge diagnoses which may have misclassified patients. Patients may have been suspected of having one diagnosis in the ED and ultimately may have had an alternative diagnosis. As such, we can only assess relative performance of the various models, but cannot draw conclusions about absolute accuracy.

This study occurred at a single institution with a custom built information system. These results might not generalize to other systems that may not be modified to support complete electronic capture of clinical data and customized decision support. 
While we internally validated the results, external validation is warranted.  It will be interesting to discover whether the same algorithm may be applied to another institution, or whether reliable machine learning requires first training on local clinical data. 

Our held-out tag prediction task is a synthetic task intended to simulate real clinical scenarios where some, but not all conditions are known about a patient. The evaluation is performed using retrospective data. Further study would be needed to confirm these results in real clinical practice.

We depend on estimates of the parameters of the corruption process to perform the method-of-moments initialization in Section~\ref{sec:init}. This is a potential weakness of the algorithm, though we are careful to consider baselines which can make use of the same information. 
In this work we do not address how those parameters could be obtained and rely on oracle estimates of these parameters. 
However, we expect that even by labeling a small number of examples we could obtain good enough estimates of these parameters to serve for the initialization. 

Diseases are not truly independent of one another, despite our modeling them as such in this paper. More elaborate method-of-moment estimation techniques can be used together with anchors to learn the joint distribution of the conditions, even when they are never observed in the data \citep{HalpernEtAl_arxiv15}.




%% file: appendix.tex
\section{Text processing}
\label{app:text}
We apply negation detection to the free-text section using ``negex'' rules \citep{Chapman2001301} with some manual adaptations appropriate for Emergency department notes \citep{JerniteEtAl_nips13health}, and replace common bigrams with a single token (e.g. ``chest pain'' is replaced by ``chest\_pain''.
the patient record). Negated terms are then added as additional vocabulary words.

The following words are indicators of the beginning of a negation: {\tt no, not, denies, without, non, unable}.

The token {\tt -} is treated as a special negation word whose scope only includes the word that follows it.

The following tokens stop a negation: 
{\tt . ; [ - newline + but and pt except reports alert complains has states secondary per did aox3}.

\section{Parameter settings for learning with Neural Variational Inference and Learning \citep{mnih2014neural}}
\label{app:parameters}
We use the signal-centering and normalization described in the paper as well as input-dependent baselines.
The input dependent baselines use a two-layer neural network with 100 hidden units and tanh activations. 
Learning rate is set to 0.0001. 10 samples are used to estimate each gradient.
When $\theta$ is initialized with method of moments, we have 50 epochs of ``burn-in'' where $\theta$ is held fixed and only $\phi$ is optimized.
The learning rate of $\theta$ is set to 1/5th of that of $\phi$.
$\pi$ parameters and the failure probabilities of the anchors are initialized using the estimated  noise rates and never optimized. 
The code is implemented in Torch and and RMSprop is used for optimization.
Failure probabilities in $\theta$ are mapped using a sigmoid function (i.e., $f_{i,j} = \sigma(\theta_{i,j})$) to allow for continuous optimization over an unbounded space.
We experimented with different values of L2 regularization (weight decay) in the recognition model, in the range \{0,0.1, 0.01, 0.001\} and chose the value that gave the best heldout-anchor performance.
Parameters of $\phi$ are initialized uniformly at random between [-0.1, 0.1].


\section{Semi-supervised objective details}
\label{app:semisupervised}
In this appendix, we expand the objective which was written in compressed notation in Section~\ref{sec:semisupervised}.

The parameters to be optimized are:
\begin{itemize}
\item $\theta \in [0, 1]^{n \times m + n + m}$: Parameters of the generative model, consisting of $n \times m$ failure probabilities ($f_{i,j}$), $m$ prior probabilities ($\pi_i$), and $n$ leak probabilities ($l_j$).
\item $\phi \in \mathbb{R}^{m \times (n+1)}$: Parameters of $m$ independent logistic regression models, one for each tag.
\item $\phi_0^\prime \in \mathbb{R}^{m}$: Additional bias terms that are introduced to account for the difference between the predictions of the recognition model, which predicts the tags, and the desired semi-supervised objective which predicts the {\em anchors}.
\end{itemize}

For the a single data point, $x$ is the binary feature vector. In practice, we center the inputs to the $q$ model (the $P$ model cannot support centering since the generative model is defined for binary variables) and pad with a single 1 to allow for a bias term. $\overline{x}$ is a centered and padded copy of $x$. $\tilde{x}$ is a copy of $\overline{x}$, with the values of the anchors set to 0. $a$ is a vector containing the binary values of the anchors from $x$. 

The likelihood of the generative model is then:
\begin{multline}
\label{eq:noisyorfull}
P(x,y; \theta) \equiv \\
\left(\prod_{i=1}^m \pi_i^{y_i}(1-\pi_i)^{1-y_i}\right) \prod_{j=1}^n \left(x_j\left(1-(1-l_j)\prod_i f_{i,j}^{y_i}\right) + (1-x_j)(1-l_j)\prod_{i=1}^m f_{i,j}^{y_i} \right)
\end{multline}

The recognition model consists of $m$ independent logistic regression models.
\begin{equation}
q(y|\overline{x}; \phi) = \prod_{i=1}^m \left( y_i \sigma(\phi_{i} \cdot \overline{x}) + (1-y_i)(1-\sigma(\phi_{i} \cdot \overline{x} ))\right),
\end{equation}
where $\sigma$ is the sigmoid function, $\sigma(x) = \frac{1}{1+e^{-x}}$.

Let $\lambda>0$ be a hyperparameter specifying the trade-off between
the lower bound on the likelihood and the semi-supervised objective
term. The final objective, for a collection of $N$ patient records,
indexed by $p$, is to maximize:
\begin{multline}
\sum_{p=1}^N \Big( E_{y \sim q(y|\overline{x}^{(p)})}\left[\log
    P(x^{(p)},y; \theta) - \log q(y|\overline{x}^{(p)})\right] +\\
  \lambda \sum_{i=1}^m\Big[ a^{(p)}_i\log\sigma(\phi_i \cdot \tilde{x}^{(p)} + \phi_{0i}^\prime) + 
  (1-a^{(p)}_i)\log(1-\sigma(\phi_i \cdot \tilde{x}^{(p)} + \phi_{0i}^\prime))\Big]\Big).
\end{multline}

\section{Leak probabilities}
\label{app:leak}
Leak probabilities are calculated to account for the difference between the actual observed counts and those predicted by the model.
If all of the failure probabilities are known, then the marginal probability of an observation predicted by the model can be calculated with the Quickscore equation~\citep{Heckerman_Quickscore}.
\begin{equation}
P(X_j=0) = (1-l_j)\prod_i (1-\pi_i + \pi_i f_{i,j}).
\end{equation}

Thus, we can solve:
\begin{equation}
\hat{l}_j =  1-\frac{\hat{P}(X_j=0)}{\prod_i (1-\pi_i + \pi_i \hat{f}_{i,j})}.
\end{equation}

This assumes that we have estimates for $\pi_i$, but that is no different from assuming that we have estimates of the noise rates $P(Y_i | A_i)$, since $\pi_i = \sum_{a_i} P(a_i) P(Y_i=1 | a_i)$, where $P(a_i)$ can be estimated from counts.

\section{Explicit matrix version of Equation~\ref{eq:matrix}}
\label{app:matrix}

{\small
\begin{eqnarray}
\label{eq:Explicit-matrix}
\vec{P}(X_j | A_i) &= \begin{bmatrix} {P}(X_j=0 | A_i=0) \\ {P}(X_j=0 | A_i=1) \\ {P}(X_j=1 | A_i=0) \\ {P}(X_j=1 | A_i=1) \end{bmatrix} \\
R &= \begin{bmatrix} P(A_i=0 | Y_i=0) & P(A_i=0 | Y_i=1) & 0 &  0  \\
                P(A_i=1 | Y_i=0) & P(A_i=1 | Y_i=0) &  & 0 \\ 
                0 & 0 & P(A_i=0 | Y_i=0) &P(A_i=0 | Y_i=1) \\ 
                0 & 0 & P(A_i=1 | Y_i=0) & P(A_i=1 | Y_i=0) 
\end{bmatrix} \\
\vec{P}(X_j | Y_i) &=\begin{bmatrix} {P}(X_j=0 | Y_i=0) \\ {P}(X_j=0 | Y_i=1) \\ {P}(X_j=1 | Y_i=0) \\ {P}(X_j=1 | Y_i=1) \end{bmatrix}.\\ \nonumber
\end{eqnarray}
}

\section{Held out anchor inference}
\label{app:lastanchor}
To perform the held out anchor inference (described in Section~\ref{sec:model_selection}), we use the fact that conditioned on some feature vector $X^\prime$ (in this case, some of the anchors are held out so it is not the full vector $X$), we have:
\begin{align*}
P(A_i| X^\prime) &= \sum_{y_i \in \{0,1\}} P(y_i | X^\prime ) P(A_i | Y_i=y_i, X^\prime) \\
&= \sum_{y_i \in \{0,1\}} P(y_i | X^\prime ) P(A_i | Y_i=y_i).
\end{align*}
We assume that the corruption rate, $P(A_i | Y_i)$, is known, and estimate $P(y_i | X^\prime)$ with Gibbs sampling.

\section{Held out tag inference}
\label{app:lasttag}
Exact inference is possible in the held out tag prediction task.
For a single data point, we have a feature vector $X$. Assume without loss of generality that the first $k$ tags known to be positive, $y_{1} = ... = y_{k} = 1$ and the final $m-k$ tags are unknown. Let $U$ refer to the unknown tags.

Of the unknown tags, we know that by the design of the task, only one is on and the rest are off. We can condition on the sum of all of the $Y$ variables being $k+1$, and we know that $U$ has to be an indicator vector (i.e., one index on and the rest off). 

We would like to calculate the likelihood that $U_i = 1$ (i.e., the ith unknown tag is on). Let $U_{-i}$ be all of the unknown tags other than $U_i$.  The likelihood is calculated as follows:

\begin{align*}
P(U_i = 1 | X, Y_{1:k} = 1, \sum y = k+1) &=  P(U_i = 1, U_{-1} = 0| X, Y_{1:k} = 1, \sum y = k+1)\\
&= \frac{P(X, U_i = 1, U_{-1} = 0,  Y_{1:k} = 1, \sum y = k+1)}{P(X, Y_{1:k} = 1 | \sum y = k+1)}\\
&= \frac{P(X, U_i = 1, U_{-1} = 0,  Y_{1:k} = 1, \sum y = k+1)}{\sum_{u \in |U|} P(X, Y_{1:k} = 1, u | \sum y = k+1)}.\\
\end{align*}

The first equality uses the fact that $U$ is an indicator vector. The second equality is from the definition of conditioning. The third line marginalizes over unknown values of $u$ in the denominator. Note that the final line uses only complete likelihoods, so each term can be calculated efficiently using Equation~\ref{eq:noisyorfull}. The sum in the denominator is over the set of indicator vectors of size $m-k$, so it can be computed efficiently as well.

\section{Optimization course}
\label{app:course}
\begin{figure}[tb]
  \centering
  \includegraphics[scale=0.38]{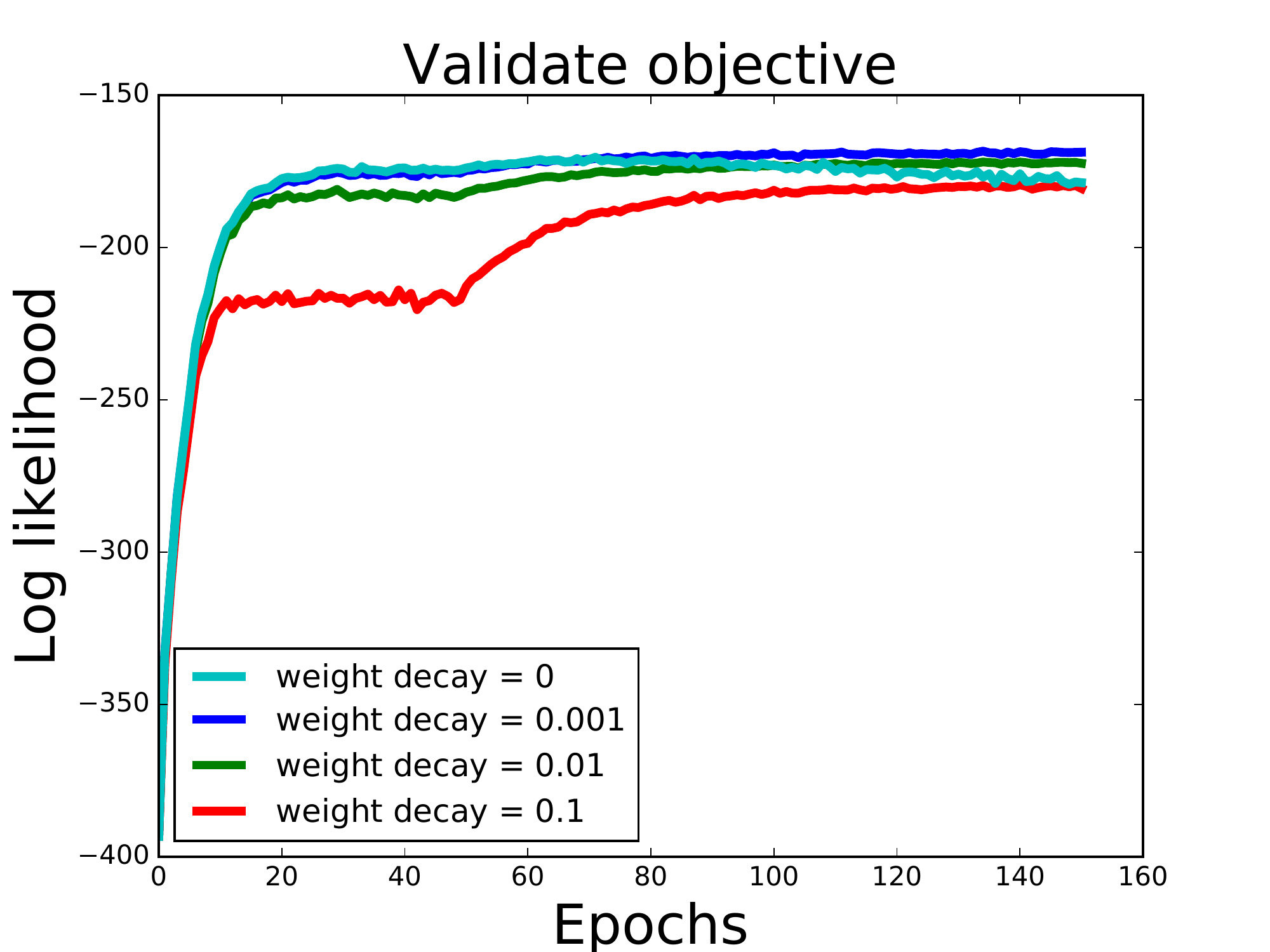}
  \includegraphics[scale=0.38]{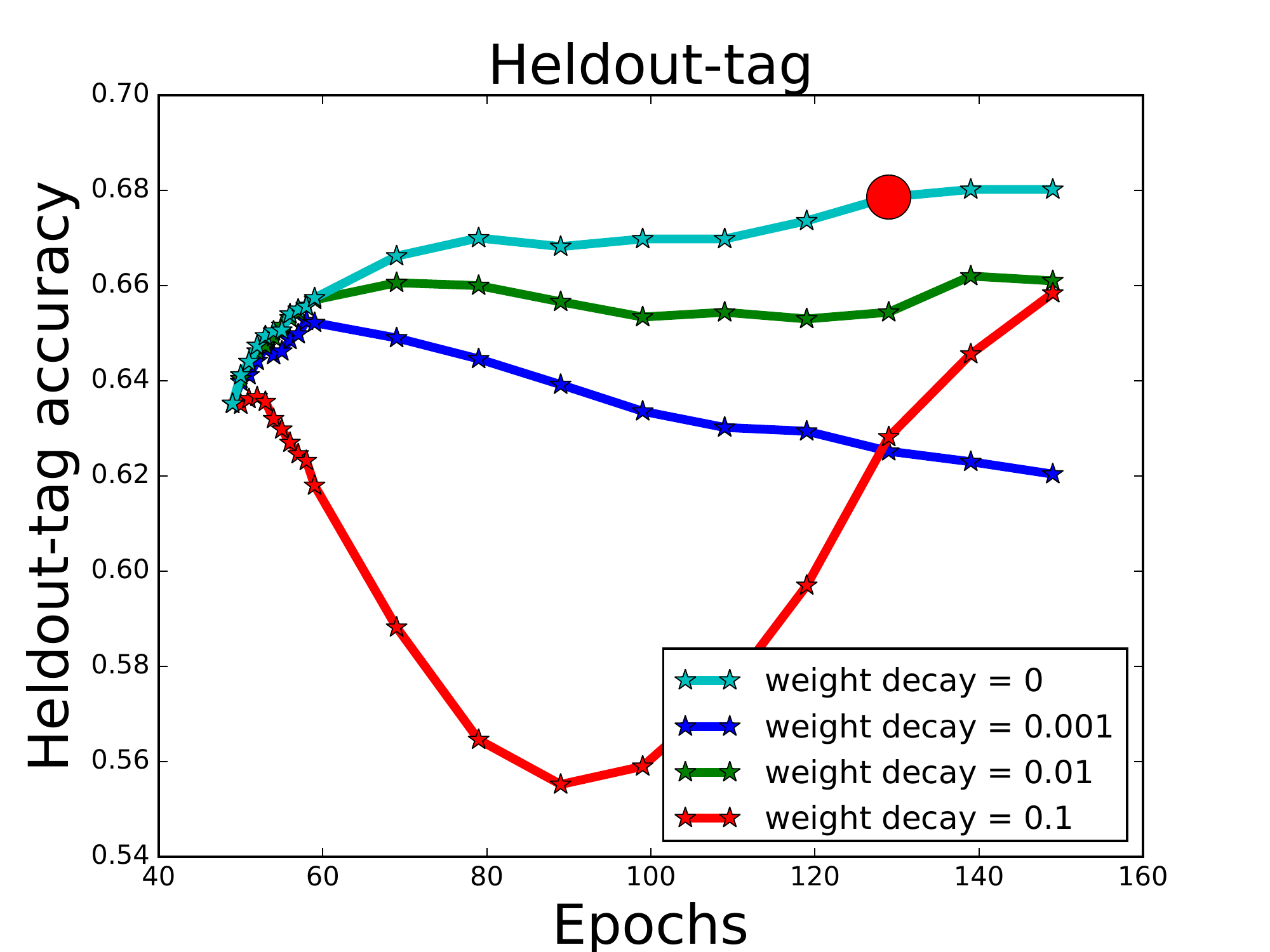}
\vspace{-7mm}
  \caption{\small Improvement of the likelihood-based objective (left) and heldout tag (right) as optimization progresses for different values of L2 regularization (weight decay) in the recognition model. The model chosen by the model selection procedure (Section~\ref{sec:model_selection}) is marked with a large red dot. The first 50 epochs are a burn-in period where only the recognition model changes.}
  \label{fig:learning}
\end{figure}

In Section~\ref{sec:semisupervised} we discuss how the likelihood objective is not aligned with the held out tag prediction task. 
Figure~\ref{fig:learning} shows that the semi-supervised objective remedies this situation, displaying how the likelihood objective and heldout tag predictions improve from the initialization baseline as optimization on the semi-supervised objective is run. 

